\documentclass[letterpaper, 10 pt, conference]{ieeeconf}  

\IEEEoverridecommandlockouts                              

\usepackage[english]{babel}
\usepackage{times} 
\usepackage{cite}
\usepackage{hyperref}
\usepackage[font=footnotesize]{subcaption}
\usepackage{calrsfs}
\usepackage[nolist]{acronym}

\DeclareMathAlphabet{\pazocal}{OMS}{zplm}{m}{n}

\usepackage{cellspace}
\newcolumntype{D}{>{\hfill}N{3}{2}<{\hfill}}

\let\labelindent\relax
\usepackage[shortlabels]{enumitem}

\usepackage{graphicx} 
\usepackage{epsfig} 
\usepackage{pgfplots}
\usepackage{caption}
\usepackage{subcaption}

\pgfplotsset{compat=1.15}
\graphicspath{{./figure/}}

\usepackage[export]{adjustbox}

\makeatletter
\let\MYcaption\@makecaption
\makeatother

\makeatletter
\let\@makecaption\MYcaption
\makeatother

\usepackage{mathptmx} 
\usepackage{amsmath} 
\usepackage{amssymb}  
\usepackage{nicefrac}
\usepackage{mathtools}
\usepackage{optidef}
\usepackage{leftidx}
\usepackage{bm} 

\makeatletter%
\AfterPreamble{%
	\usepackage{hyperref}%
	\let\oldhypertarget\hypertarget%
	\renewcommand{\hypertarget}[2]{%
		\oldhypertarget{#1}{#2}%
		\protected@write\@mainaux{}{%
			\string\expandafter\string\gdef%
			\string\csname\string\detokenize{#1}\string\endcsname{#2}%
		}%
	}%
	\newcommand{\myhyperlink}[1]{%
		\hyperlink{#1}{\csname #1\endcsname}%
	}%
}
\makeatother%

\newcounter{Definition}
\newcounter{Theorem}
\makeatletter
\providecommand{\bigsqcap}{%
	\mathop{%
		\mathpalette\@updown\bigsqcup
	}%
}
\newcommand*{\@updown}[2]{%
	\rotatebox[origin=c]{180}{$\m@th#1#2$}%
}

\makeatother

\usepackage{tikz}
\usetikzlibrary{quotes,angles,calc}

\tikzset{
    imglabel/.style={
      rectangle,
      inner sep=2pt,
      text=black,
      minimum height=1em,
      text centered,
      fill=white,
      fill opacity=1.0,
      text opacity=1,
      anchor=south west,
    },
  }

\usepackage{siunitx}
\usepackage{todonotes}
\usepackage{multirow}

\usepackage{algorithm}
\usepackage[noend]{algpseudocode}

\makeatletter
\def\BState{\State\hskip-\ALG@thistlm}
\makeatother

\newcommand\copyrighttext{%
	\small \begin{center} \color{red} \textcopyright\,2021 IEEE. Personal use of this material is 	permitted. Permission from IEEE must be obtained for all other uses, in any current or future 	media, including reprinting/republishing this material for advertising or promotional purposes, creating new collective works, for resale or redistribution to servers or lists, or reuse 
	of any copyrighted component of this work in other works. \end{center}}
\newcommand\copyrightnotice{%
	\begin{tikzpicture}[remember picture,overlay]
	\node[anchor=south,yshift=25.6cm] at (current page.south) 
	{\color{red}\fbox{\parbox{\dimexpr\textwidth-\fboxsep-\fboxrule\relax}{\copyrighttext}}};
	\end{tikzpicture}%
}

\title{\copyrightnotice \LARGE \bf
An Application of Stereo Thermal Vision for Preliminary Inspection of Electrical Power Lines by MAVs
}

\author{Lyubomyr Demkiv$^1$, Massimiliano Ruffo$^2$, Giuseppe Silano$^3$, Jan Bednar$^3$, and Martin Saska$^3$
	\thanks{$^1$Lyubomyr Demkiv is with the Lviv Polytechnic National University, Ukraine (email:  {\tt\small lyubomyr.i.demkiv@lpnu.ua)}.}
	\thanks{$^2$Massimiliano  Ruffo is with Terabee, Switzerland (email: {\tt\small max.ruffo@terabee.com)}.}
	\thanks{$^3$Giuseppe Silano, Jan Bednar, and Martin Saska are with the Faculty of Electrical Engineering, Czech Technical University in Prague, Czech Republic (email: {\tt\small \{giuseppe.silano, jan.bednar14, martin.saska\}@fel.cvut.cz)}.}
	\thanks{This work was partially funded by the European Union's Horizon 2020 research and 
	innovation programme AERIAL-CORE under grant agreement no. 871479, by Czech Science Foundation
	(GAČR) under research project no. 20-10280S, and by TAČR within research project no. 
	FW01010317.}
}

\begin{document}
	
\maketitle
\thispagestyle{empty}
\pagestyle{empty}


\begin{abstract}
	
An application of stereo thermal vision to perform preliminary inspection operations of electrical power lines by a particular class of small~\acp{UAV}, aka~\acp{MAV}, is presented in this paper. The proposed hardware and software setup allows the detection of overheated power equipment, one of the major causes of power outages. The stereo vision complements the GPS information by finely detecting the potential source of damage while also providing a measure of the harm extension. The reduced sizes and the light weight of the vehicle enable to survey areas otherwise difficult to access with standard~\acp{UAV}. Gazebo simulations and real flight experiments demonstrate the feasibility and effectiveness of the proposed setup.

\end{abstract}



\begin{acronym}
	\acro{CNN}[CNN]{Convolutional Neural Network}
	\acro{FOV}[FoV]{Field of View}
	\acro{FSM}[FSM]{Failure recovery and Synchronization jobs Manager}
	\acro{ICP}[ICP]{Iterative Closest Point}
	\acro{IR}[IR]{Infrared}
	\acro{GNSS}[GNSS]{Global Navigation Satellite System}
	\acro{GPS}[GPS]{Global Positioning System}
	\acro{MAV}[MAV]{Micro Unmanned Aerial Vehicle}
	\acro{MBZIRC}[MBZIRC 2020]{Mohamed Bin Zayed International Robotics Challenge 2020}
	\acro{MBZIRC17}[MBZIRC 2017]{Mohamed Bin Zayed International Robotics Challenge 2017}
	\acro{MIDGARD}[MIDGARD]{MAV Identification Dataset Generated Automatically in Real-world 
	Deployment}
	\acro{MOCAP}[mo-cap]{Motion Capture}
	\acro{MPC}[MPC]{Model Predictive Control}
	\acro{MRS}[MRS]{Multi-Robot System}
	\acro{ML}[ML]{Machine Learning}
	\acro{NN}[NN]{Neural Network}
	\acro{ROS}[ROS]{Robot Operating System}
	\acro{ROW}[ROW]{Rolling On Wire}
	\acro{RTK}[RTK]{Real-time Kinematic}
	\acro{SIL}[SIL]{Software-in-the-loop}
	\acro{UAV}[UAV]{Unmanned Aerial Vehicle}
	\acro{UGV}[UGV]{Unmanned Ground Vehicle}
	\acro{UV}[UV]{UltraViolet}
	\acro{UVDAR}[\emph{UVDAR}]{UltraViolet Direction And Ranging}
	\acro{UT}[UT]{Unscented Transform}
	\acro{RMSE}[RMSE]{Root Mean Square Error}
	\acro{wrt}[w.r.t.]{with respect to}
\end{acronym}






\section{Introduction}
\label{sec:introduction}


Over the last two decades, the global energy demand has increased rapidly due to demographic and economic growth, especially in emerging market areas. This has created new challenges for electricity supply companies, which are constantly looking for new solutions to minimize the frequency of power outages. Power failures are particularly critical when the environment and public safety are at risk, e.g., for hospitals, sewage treatment plants, and telecommunication systems. One of the major causes of a power outage is damage to transmission lines, usually due to storms, overheating phenomena, or inefficient inspection campaigns~\cite{EPRI2011TechnicalReport, Park2019JFR, AhajjamIWCMC2020}.

\begin{figure}
	\begin{center}
		\adjincludegraphics[width=0.45\textwidth, trim={{0.0\width} {.15\height} {0.0\width} {.10\height}}, clip]{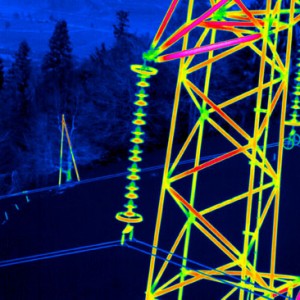}
	\end{center}
	\caption{A false color image of a power tower. Colors provide thermal analysis of power equipment by highlighting the presence of overheated objects.}
	\label{fig:exampleIRInspection}
\end{figure}

Nowadays, the most common strategy for reducing energy interruptions is to schedule periodic 
maintenance activities by carrying out repairs and replacements on active lines~\cite{LIU2020ARC, Matikainen2016ISPRS}. Manned helicopters equipped with thermal, infrared, and ultraviolet light cameras and experienced crews assess the operating condition of power equipment (e.g., switchboards, conductors, insulators) detecting potential problems and faults, such as high resistance contacts, inductive heating, short and open circuits (see Fig.~\ref{fig:exampleIRInspection}). However, there are two major drawbacks to this approach: first, inspection and maintenance are dangerous for operators who have to fly close to power towers and operate on electrified lines; second, the operations are extremely time-consuming and expensive (\$1,500 for a one-hour flight) and prone to human error~\cite{Baik2018JIRS, Martinez2018EAAI}. 

Therefore, there is a clear need for safe and practical setups that enable more efficient maintenance and inspection procedures in electrical power lines, thereby reducing risks and costs for distribution companies. Multiple solutions have been investigated in the literature for automating this task~\cite{LIU2020ARC}, but the most promising and flexible option is to use~\acfp{UAV}, as they are capable of supporting inspection at different levels~\cite{Martinez2018EAAI, Silano2021EUSIPCO}. For instance,~\acp{UAV} can monitor places of difficult access making the inspection operations faster and more effective.

\begin{figure}
    \centering
    \includegraphics[width=0.45\textwidth, keepaspectratio]{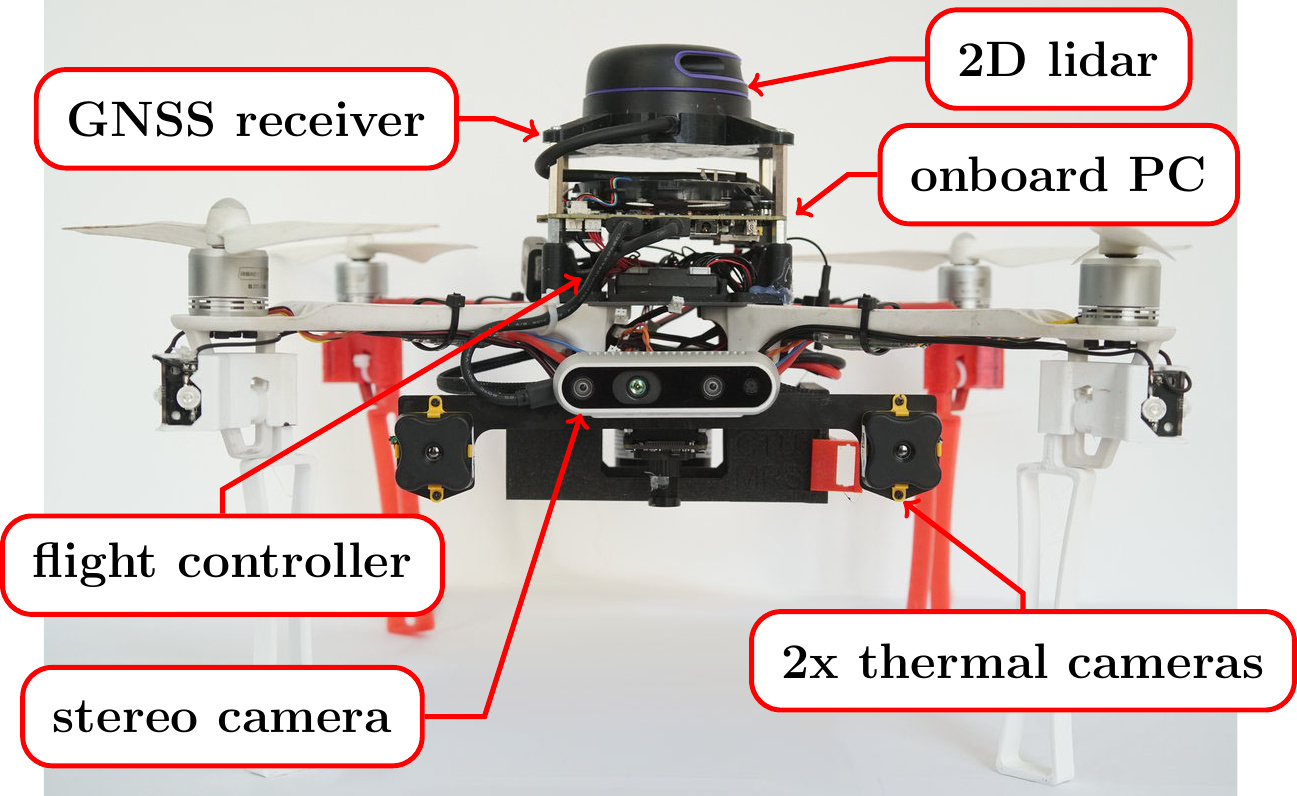}
    \caption{The proposed~\ac{MAV} platform along with a description of the sensory equipment layout.}
    \label{fig:mavPlatform}
\end{figure}

However, the use of~\acp{UAV} to achieve these tasks is particularly challenging, due to their limited battery capacity, the strong electromagnetic interference produced by power towers, and the presence of potential obstacles along the lines (e.g., branches, vegetation, marker balls)~\cite{Silano2021ICUAS}. Small and lightweight systems with cognitive capabilities, e.g., based on novel perception sensors~\cite{WalterRAL2019}, advanced data fusion techniques~\cite{Baca2020mrs}, planning~\cite{SilanoRAL2021, Nekovar2021RAL} and control algorithms~\cite{KratkyRAL2021}, are of interest to address those complexities and to accomplish the assigned mission safely and successfully. More precisely, it is key to integrate small~\acp{UAV} (see Fig.~\ref{fig:mavPlatform}), aka~\acfp{MAV}, capable of detecting overheated power equipment and recovering their relative position~\ac{wrt} the vehicle by complementing the GPS information. In addition, such actions should be performed to ensure compliance with safety requirements, i.e., maintaining a safe distance from the tower, avoiding collisions with the power equipment, possibly reacting to unforeseen events and external disturbances, during the course of the mission.

Versatile and reliable hardware and software architectures are essential to integrate these features, especially when many interconnected and heterogeneous components (e.g., path planners, control and computer vision algorithms, data transmission systems) work together. Therefore, in this paper, we propose such a setup to deal with the detection and finely location of potential source of damage in the context of power line inspection.

\subsection{Related works}
\label{sec:relatedWorks}

\begin{figure}
  \centering
  \includegraphics[trim={0pt 0pt 0pt 0pt},clip,width=0.98\columnwidth]{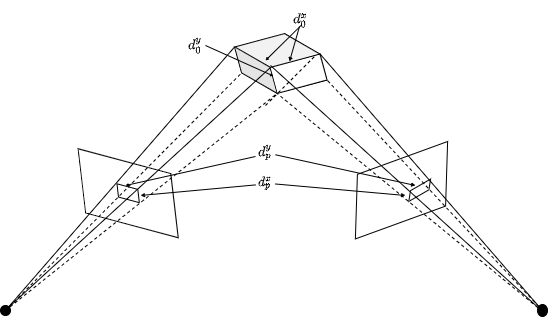}
  \caption{Illustration of the epipolar geometry. The two cameras are indicated by their centres as two black points and image planes. An image point back-projects to a ray in 3D space. The width and height of the object in the camera ($d_p^x, d_p^y$) and 3-dimensional scene ($d_0^x, d_0^y$) are also reported.}
  \label{fig:steadyZone}
\end{figure}

Thermal vision has been extensively studied in recent years as valuable tool for inspection and surveillance purposes~\cite{Wend2108ICMV, Kim2021RAL}, soil and field analysis and crop monitoring~\cite{Swati2021CCT}. However, there are two main challenges for~\acp{UAV} working on stereo thermal vision: (i) \textit{the correspondence problem}, to ensure that the point on one image matches its counterpart on the other (see Fig.~\ref{fig:steadyZone}); (ii) \textit{reconstruction}, to recover 3D points from known point matches. 

Much of the state-of-the-art focuses on the first problem. Some works~\cite{Dorit2012IFAC, Lauterbach2019SSRR} propose to relax the correspondence constraints by exploiting the epipolar geometry and using the \textit{pinhole} camera model. Roughly speaking, given a pair of calibrated cameras, for a point in the left (right) image, the set of possible matches in the right (left) image is constrained to lie on the corresponding epipolar line, in order to reduce the search space from the whole image to a line. Others~\cite{Prakash2006CARV, Vidas2013ICRA} deal with recovering local measurements, such as image intensity and phase, and aggregating information from multiple pixels by using smoothness constraints. In this case, very simple aggregation algorithms can be used, such as the Kanade matching error algorithm detailed in~\cite{Okutomi1993TPAMI}. 

As regards the reconstruction problem, several approaches focus on the rectification of un-calibrated stereo images and real-time depth estimation and measurement of objects~\cite{Starr2013ICAIM, Stojcsics2018INES}. Various solutions have also been proposed for finding the distance of moving objects based on color and feature matching~\cite{Stojcsics2018INES}. However, most of these problems rely on the use of high-definition thermal cameras ($1$ megapixel). Although this resolution is a long way from that of standard RGB cameras (typically around $16$ megapixels), the prices themselves, sizes and weight make them difficult to use on embedded system, particularly on aerial vehicles. Moreover, even when the hardware specifications become reasonable for the application, the main weakness of the algorithms lies in high computation times which are impractical for inspection operations.



\subsection{Contributions}
\label{sec:contributions}

In this paper, we propose an application of stereo thermal vision for preliminary inspection operations of electrical power lines and their power equipment. A \ac{MAV} equipped with small sizes, lightweight, and economical thermal cameras is used to cope with the task. The described hardware and software setup is designed for the AERIAL-CORE European project\footnote{\url{https://aerial-core.eu}} and is built on top the~\ac{ROS} system~\cite{Baca2020mrs}. The Nimbro network\footnote{\url{https://github.com/ctu-mrs/nimbro_network}} supports the communication between the vehicle and a base station by implementing the \textit{TCP Fast-Open protocol} and reducing bandwith using the \textit{libbz2} data compression algorithm. The framework combines the advantages of stereo vision for depth estimation and object reconstruction with that of thermal imaging for a non-invasive and detailed inspection of electrical components even in poor lighting conditions.

The advantages are twofold: (i) the use of low-resolution cameras speeds up the inspection operations by reducing the number of pixels, thus allowing a reduction in computation times. In fact, the more pixels there are, the longer the algorithm takes to process the data; (ii) the reduced sizes and the light weight of the aerial vehicle enable to survey areas of difficult access and dangerous for the operators, such as the supports for bare conductors, while also providing a measure of the harm extension.

Gazebo simulations and real flight experiments showcase the capabilities and potential of the platform. Moreover, they demonstrate an advanced level of reliability of the whole inspection system in the AERIAL-CORE project.







\section{System Overview}
\label{sec:systemOverview}

The proposed hardware platform is based on the DJI F450 quad-rotor, equipped with an Intel NUC onboard computer (an i7-8559U processor with $16$GB of RAM) and the Pixhawk flight controller. The software stack is embedded by using the Melodic Morenia version of~\ac{ROS} running on Ubuntu $18.04$. The code is released as open-source and available at~\url{https://github.com/ctu-mrs/mrs_uav_system}. 

The~\ac{MAV} is also equipped with a GNSS module along with a Ublox Neo-M8N receiver. The sensory equipment of the vehicle includes a RPLIDAR rotary rangefinder for 2D localization, a Garmin laser rangefinder used as altimeter, a BlueFOX-MLC200wC camera to recover the optical flow from the ground, and two Teraranger EVO thermal\footnote{\url{https://www.terabee.com/shop/thermal-cameras/teraranger-evo-thermal-33}} cameras with a~\ac{FOV} of $\SI{33}{\degree} \times \SI{33}{\degree}$ as for stereo thermal vision. The two thermal cameras are mounted in compliance with 3D localization considerations taking into account their resolution and the operating conditions, as described in Section~\ref{sec:stereoGeometry}. 

The control architecture and the reactive obstacle avoidance and localization modules are based on~\cite{Baca2020mrs, Petrlik2021ICUAS, Silano2021ICUAS}. Figure~\ref{fig:mavPlatform} shows the sensory equipment layout onboard the~\ac{MAV} platform.



\subsection{Vision}
\label{sec:vision}

The thermal cameras provide as output a matrix of $32 \times 32$ pixels, where the value of a pixel represents the intensity of the~\ac{IR} thermal radiation emitted by the object. Then, the~\ac{IR} intensity can be converted into the corresponding temperature level expressed in degrees Celsius using the Stefan-Boltzmann equation and the surface emissivity value~\cite{Spurny2021Access}. This value is a measure of the efficiency in which a surface emits thermal energy and ranges from $0$ to $1$.

For most surfaces (e.g., plastic, iron, glass), the emissivity is approximately $0.95$, which is the value used by the selected sensors \cite{evo_thermal}. However, this value can differ greatly for certain materials such as aluminum, leading to lower temperature readings. In fact, it is worth noticing that if the material is highly reflective in the given~\ac{IR} range, the camera's output in that region will represent the properties of the reflected surfaces instead of the reflective material itself, akin to a mirror. 

For all such reasons, the proposed setup is meant as a preliminary inspection tool for electrical power lines. In all the performed experiments, overheated objects with a temperature above than $\SI{125}{\celsius}$ were considered. In this case, the thermal cameras would be sufficient to detect overheating phenomena from a "significant" distance with good reliability and without running into outlier detection.

\section{Image processing}
\label{sec:imageProcessing}

The pair of images from thermal cameras are processed in multiple steps. These can be divided into three phases: first, the centroids of individual enclosed features are retrieved from both images; second, these pixel positions are matched with their pairs in between the two images; and third, from the disparity\footnote{The difference in distance between pixel positions of the target in the right and left image.} in their image positions, the relative 3D positions of the observed targets are calculated.




\subsection{Target object segmentation and matching}
\label{sec:segemntation}


In order to obtain from the thermal cameras information suitable for relative pose estimation, the following steps are performed:

\begin{enumerate}[1)]

    \item \textit{Retrieve images from left and right thermal camera}. Images from the two cameras arrive asynchronously. Therefore, in order to minimize parasitic effects of motion on the distance estimate, camera frames from each camera are selected in such a way that the difference in their individual acquisition time is minimal.
    
    \item \textit{Filter the images to reduce noise}. Values outside a selected temperature range are set to a specific value interpreted as the background. A heuristic approach based on acquired experimental dataset is used to set these thresholds by finely detecting the potential source of damage.

    \item \textit{Detect the centers of heat objects}. For the purposes of validating the proposed approach, it is assumed that the target object is a roughly circular blob.
    The corresponding pixel blobs not marked as a background are interpreted as the contours of the heated objects of interest. The centroid of these pixels is then taken as representative image coordinates of the object. Additionally, the average temperature of the blob is stored for further disambiguation.
    
  
    
\end{enumerate}

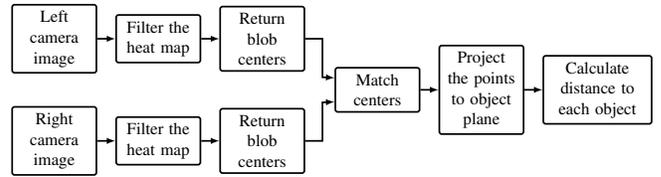
\begin{figure}
  \centering
  \resizebox{\columnwidth}{!}{
    \begin{tikzpicture}[>=latex, line width =0.5mm, minimum height=.5em, text centered,
      boxed/.style={%
        draw,
        rounded corners=0.2em,
        inner sep = .5em,
        text width = 10em,
        minimum height = 3.4em,
        text centered,
        font ={ \Large}
      },
      bubbled/.style={%
        draw,
        dotted,
        rounded corners=1.0em,
        inner sep=.7em,
      }
      ]
      \node[anchor=south west,inner sep = .01em,] (a) at (0,0) {
          \def\arraystretch{0}
        };
      \begin{scope}[x={(a.south east)},y={(a.north west)}]
        \node[boxed, text width = 8em](distance) at (0.0,0.0) {{Calculate distance to each object}};
        \node[boxed,left= 1.5em of distance, text width = 6em](project) {{Project the points to object plane}};
        \node[boxed,left= 1.5em of project, text width = 6em](match) {{Match centers}};
        
        \coordinate (anchor1) at ($ (match.west) + (-6em,0em)$);

        \node[boxed,above= 1.5em of anchor1, text width = 6em](l_blob) {{Return blob centers}};
        \node[boxed,below= 1.5em of anchor1, text width = 6em](r_blob) {{Return blob centers}};
        \node[boxed,left= 1.5em of l_blob, text width = 6em](l_filter) {{Filter the heat map}};
        \node[boxed,left= 1.5em of r_blob, text width = 6em](r_filter) {{Filter the heat map}};
        \node[boxed,left= 1.5em of l_filter, text width = 6em](l_camera) {{Left camera image}};
        \node[boxed,left= 1.5em of r_filter, text width = 6em](r_camera) {{Right camera image}};

        \draw[draw=black,->] (project) -- (distance);
        \draw[draw=black,->] (match) -- (project);
        \draw[draw=black,->] (l_blob) -- ++(5em,0em) -| ++(0em,-3.2em) -- ($ (match.west) + (0em,1.0em)$);
        \draw[draw=black,->] (r_blob) -- ++(5em,0em) -| ++(0em,3.2em) -- ($ (match.west) + (0em,-1em)$);
        \draw[draw=black,->] (l_filter) -- (l_blob);
        \draw[draw=black,->] (r_filter) -- (r_blob);
        \draw[draw=black,->] (l_camera) -- (l_filter);
        \draw[draw=black,->] (r_camera) -- (r_filter);

      \end{scope}
    \end{tikzpicture}
  }
  \caption{Flow chart of the depth estimation algorithm. Left and right images are combined to retrieve the distance to each object. 
  }
  \label{fig:flowchart_algorithm}
\end{figure}

The flow chart of the algorithm is shown in Fig.~\ref{fig:flowchart_algorithm}.

After the centroids of the pixel blobs (i.e., heat centers) are obtained for each camera individually, the centers from both cameras need to be matched into pairs to calculate the disparity and thus the depth of the heat source. The point matching algorithm can be split into three steps: (i) polygon matching, (ii) polygon normalization, and (iii) coordinate systems merging for point matching. The idea of the algorithm is inspired by the Generalized Hough transform~\cite{Ballard1981}, and customized to fit the paper outcome: shape matching for the case of arbitrary polygons. In this case the goal was to find the correspondence between the points that are detected by the thermal cameras. 

\begin{figure}
  \begin{center}
  \subfloat[One point]{%
  \includegraphics[trim={0pt 0pt 0pt 0pt},clip,width=0.5\columnwidth]{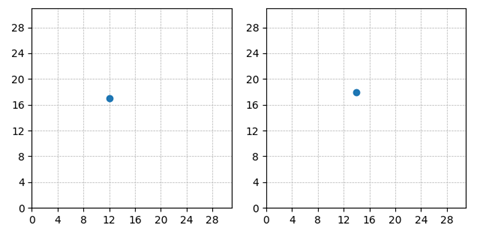}%
  \label{fig:placement_one}
  }
  \subfloat[Equal number gr. one]{%
  \includegraphics[trim={0pt 0pt 0pt 0pt},clip,width=0.5\columnwidth]{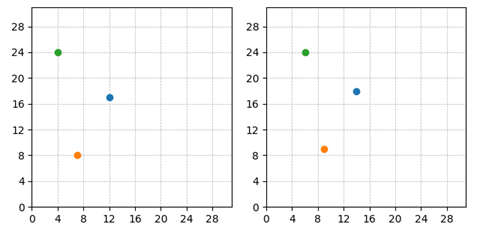}%
  \label{fig:placement_equal}
  }\\
  \subfloat[Unequal number gr. one]{%
  \includegraphics[trim={0pt 0pt 0pt 0pt},clip,width=0.5\columnwidth]{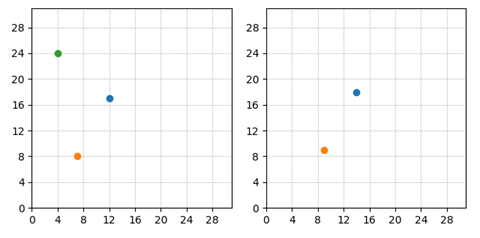}%
  \label{fig:placement_unequal}
  }
  \subfloat[No point on one of the images]{%
  \includegraphics[trim={0pt 0pt 0pt 0pt},clip,width=0.5\columnwidth]{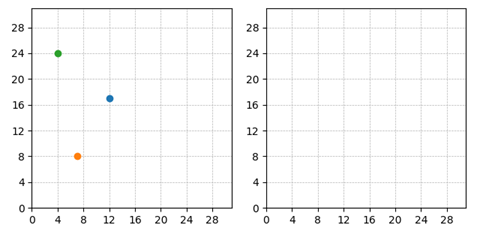}%
  \label{fig:placement_none}
  }
  \end{center}
  \caption{Possible points placement combinations. The axes are labeled in pixels. Colors represent the heat objects detected by the thermal cameras.}
  \label{fig:detectionScenarios}
\end{figure}

In the step of \textit{polygon matching}, there are the following possible states to address:

\begin{enumerate}[(A)]

  \item A single heat point is acquired from both cameras (see Fig.~\ref{fig:placement_one});\label{case_one}

  \item Equal and non-zero number of heat points is acquired from both cameras (see Fig.~\ref{fig:placement_equal});\label{case_two}
  
  \item Unequal and non-zero number of heat points is acquired from each camera (see Fig.~\ref{fig:placement_unequal});\label{case_three}
  
  \item Heat points are only acquired from one camera (see Fig.~\ref{fig:placement_none}).\label{case_four}

\end{enumerate}
 
\begin{figure}
  \begin{center}
  \subfloat[Polygon matching]{%
  \includegraphics[trim={0pt 0pt 0pt 0pt},clip,width=0.5\columnwidth]{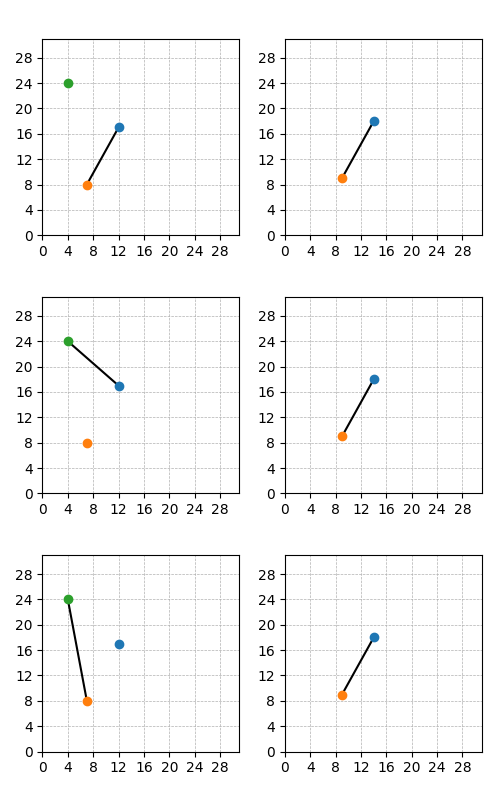}%
  \label{fig:polygon_matching}
  }
  \subfloat[Polygon normalization]{%
  \includegraphics[trim={0pt 0pt 0pt 0pt},clip,width=0.5\columnwidth]{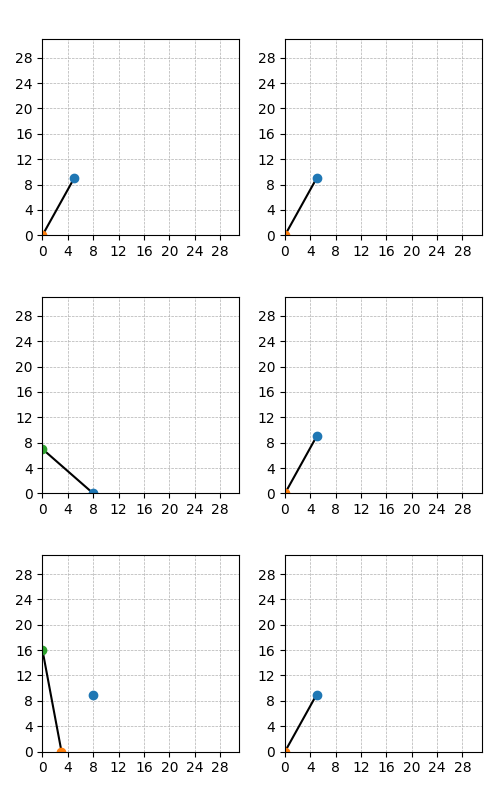}%
  \label{fig:polygon_normalization}
  }
  \end{center}
  \caption{Polygon matching and normalization in case~\ref{case_three}.}
  \label{fig:polygons}
\end{figure}

For case \ref{case_four}, it can simply skip matching. For case \ref{case_one}, it can be assumed that two detected points correspond to the same object and no further matching is needed. Lastly, in cases \ref{case_two} and \ref{case_three} the matching process starts with making all possible polygon combinations on each image (i.e., number of vertices in both polygons are the same and equal to smallest number of points on each image). When the number of points on both images are equal, i.e., there is only one possible polygon for each image (case \ref{case_two}), no further permutation is needed. Conversely, when the number of heat points from both images is not equal (case \ref{case_three}), the range of possible combinations to test for is greater, as illustrated in Fig.~\ref{fig:polygon_matching}.

\begin{figure}
  \begin{center}
  \includegraphics[trim={0pt 0pt 0pt 0pt},clip,width=0.5\columnwidth]{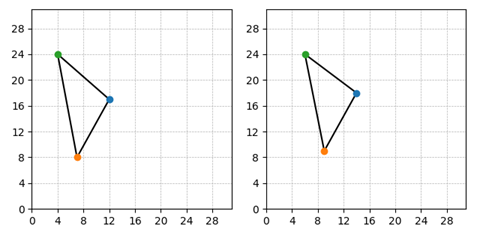}%
  \end{center}
  \caption{Polygons obtained from Fig. \ref{fig:placement_equal}.}
  \label{fig:polygon_3b}
\end{figure}

In all the scenarios depicted in Fig.~\ref{fig:detectionScenarios}, the gray line (see Fig.~\ref{fig:polygon_3b}) is meant to represent the pair of heat points on the image tested for similarity to the pair of heat points in the other image. The points that are not connected are not considered any further. Such situation might occur when the heat elements are close to the edge of~\ac{FOV} of one of the cameras and therefore might be not visible by the other camera.


After selecting each pair of candidate polygons for matching, their similarity is calculated and the pair with the best match is selected. For this purpose, the \textit{polygons are normalized} by subtracting the coordinates of all points in the image (see Fig.~\ref{fig:polygon_normalization}). Then, the Euclidean distance between each pair of ordered points for every polygon pair is found and the minimum sum of absolute distances is used to retrieve the best matching polygon. As for the point ordering, the entire procedure can be summarized via the following steps:
\begin{enumerate}[1)]
  \item Find the centroid of each polygon;
  \item Find vectors from each point of the polygon to the centroid;
  \item Find angle of each vector~\ac{wrt} the $x$- and $y$-axis of the camera frame;
  \item Match the vectors by minimizing the mutual~\ac{RMSE} of the angles in (3);
  \item Reject matches where the corresponding average temperatures differ by more than \SI{10}{\celsius}.
\end{enumerate}



It is worth noticing that compared to RGB vision, typically used in stereo vision setups, thermal vision - especially with low-resolution and unsynchronized cameras - is subject to multiple adverse effects that preclude the use of similarity-based search techniques along known epipolar lines. These effects include, among others, the splitting a heated object into two contours in one view, the \textit{thermal glare} that distorts the shape of the contours, the shift of the camera position in between acquisition times of the two images due to vibrations, reflectivity of an object that exceeds the emissivity in its effect on the image only from certain angles, and the fact that a single sub-matrix sample that can be used for reasonable comparison between the two images will cover a large portion of the low-resolution image.


Once correspondences between heat points from the two cameras have been established, their depth can be calculated by looking for \textit{coordinate systems merging for point matching}. For each point in the matched normalized polygon, the Euclidean distance between its image position and the position of the corresponding point on the other image is found. Such a distance is also known as \textit{disparity}, and it is the basis for distance estimation in stereo vision. In most cases, the offset between the image points is expected to be horizontal, given that the two cameras are aligned and placed along the horizontal image axis of each other.



\subsection{3D localization considerations}
\label{sec:stereoGeometry}



The distance estimation from a pair of images from identical \textit{pinhole-model} cameras that are aligned, or their images are rectified, is defined as follows:
\begin{equation}
  z_e = \frac{f\cdot b}{d} = \frac{f\cdot b}{x_l-x_r},
\end{equation}
%
where $z_e$ is the estimated distance of the target from the camera origins along their aligned optical axes, $f$ is the focal length in pixels, $b$ is the metric length of the baseline, and $x_r$ and $x_l$ are the horizontal pixel positions of the target in the right and left image, respectively, with their difference being called \textit{disparity} and indicated with $d$. This allows estimating the theoretical range of error in such distance calculation, given the finitely small size of a pixel. This measure is relevant for the considered scenarios due to the relatively small resolution of the cameras~\ac{wrt} their~\ac{FOV}.

\begin{figure}
  \begin{center}
  \subfloat[]{%
    \hspace{-1.75em}
    \includegraphics[width=0.52\linewidth, trim=0cm 0cm 0cm 0cm, clip=true]{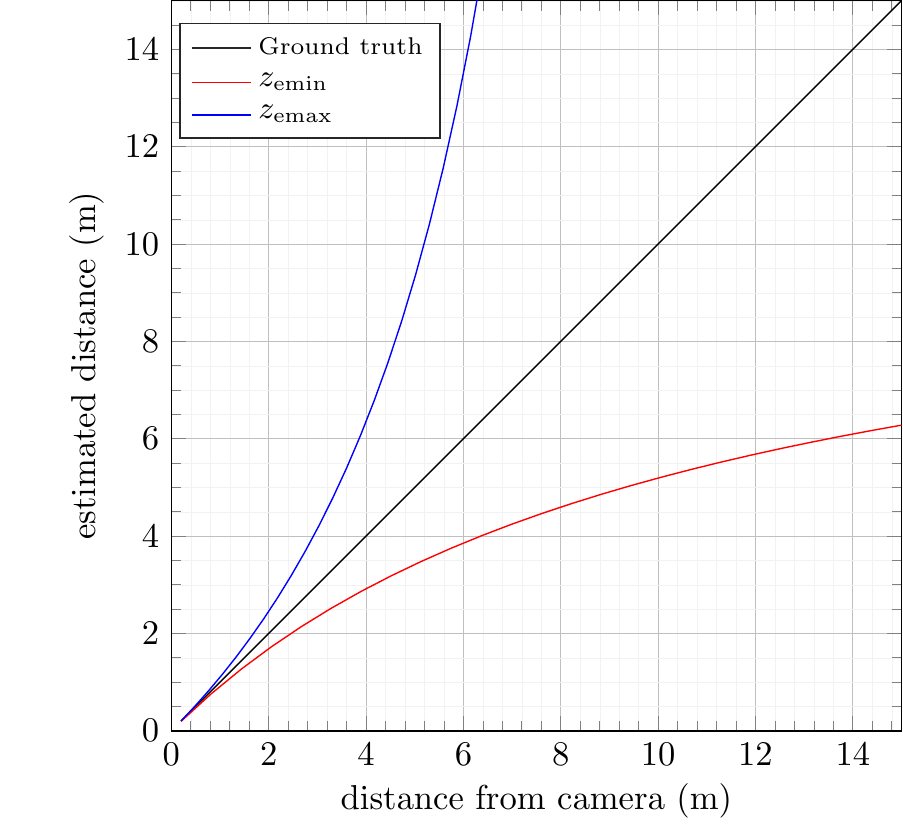}%
    \label{fig:errorOverDist}%
  }
 \hspace{0.05em}
 \subfloat[]{%
    \includegraphics[width=0.52\linewidth, trim=0cm 0cm 0cm 0cm, clip=true]{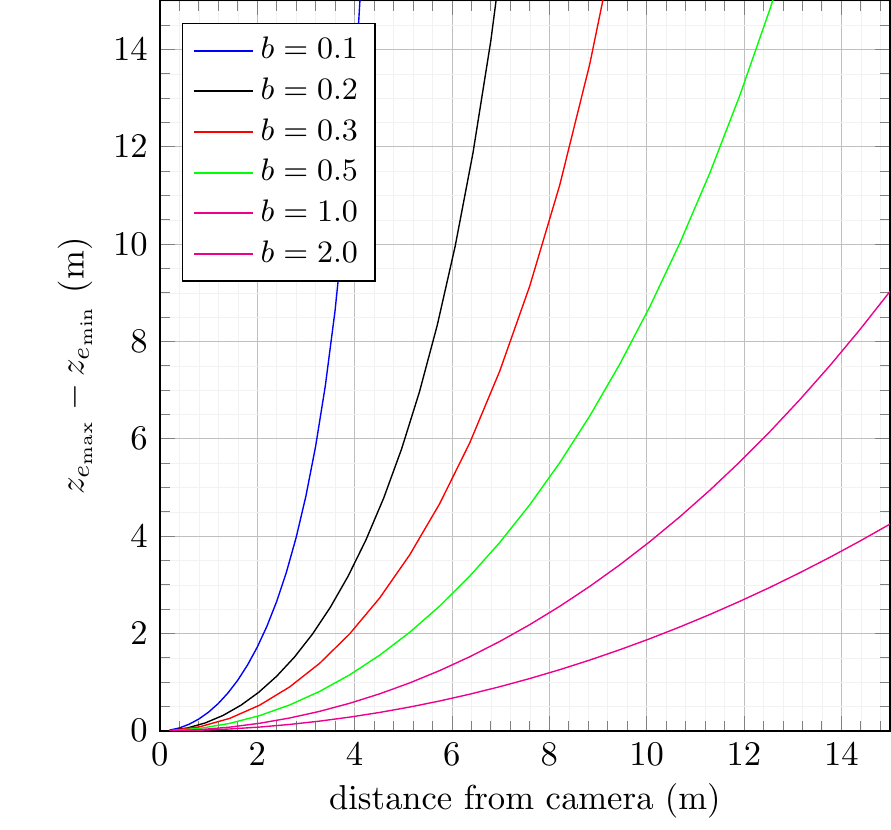}%
    \label{fig:errorRangesBaseline}%
  }
  \end{center}
  \caption{From left to right: the estimated theoretical maximum ($z_{e_\mathrm{max}}$) and minimum ($z_{e_\mathrm{min}}$) distances and the theoretical distance range $\left(z_{e_\mathrm{max}} - z_{e_\mathrm{min}}\right)$ for various values of the baseline ($b$).}%
  \label{fig:estimatedTheoreticalDistances}%
\end{figure}

Since $x_l$ and $x_r$ are usually interpreted as representing the image position in the center of their corresponding pixel, the disparity error $e_d$ caused by the finite pixel size can be $\pm 1$ pixel - the object being located on the closer edges or on the farther of these pixels. Therefore, the ideal disparity $d_t$ with perfect sub-pixel accuracy generated in the stereo pair by object at the true distance $z_t$ can be defined as:
\begin{equation}
  d_t = \frac{f\cdot b}{z_t}.
  \label{eq:disparity_theoretical}
\end{equation}
Hence, the closest distance corresponding to the imperfect measurement is:
\begin{equation}
  z_{e_\mathrm{min}} = \frac{f\cdot b}{\frac{f\cdot b}{z_t} + 1} = \frac{f\cdot b\cdot z_t}{f\cdot b + z_t},
  \label{eq:mindist}
\end{equation}
and the farthest distance is:
\begin{equation}
  z_{e_\mathrm{max}} = \frac{f\cdot b}{\frac{f\cdot b}{z_t} - 1} = \frac{f\cdot b\cdot z_t}{f\cdot b - z_t}.
  \label{eq:maxdist}
\end{equation}
Let us also define the focal length $f$ of a \textit{pinhole} camera as:
\begin{equation}\label{eq:focalLength}
  f = \frac{W/2}{\tan(\xi/2)},
\end{equation}
where $W$ is the image width in pixels and $\xi$ is the~\ac{FOV}, then the focal length for the considered Teraranger thermal camera will be $f \approx 54~\text{pixels}$. Obviously, the greater the baseline, the smaller is the influence of the pixel error on the distance estimation precision, but the greater is also the minimal measurable target distance, due to the diminishing overlap of the~\acp{FOV}, as described in Fig.~\ref{fig:estimatedTheoreticalDistances}. Additionally, extreme baselines would lead to issues with construction rigidity and carrying capacity of the~\ac{MAV}.



\begin{figure}
    \hspace{-1.5em}
    \centering
    \includegraphics[width=0.93\linewidth, trim=0cm 0cm 0cm 0cm, clip=true]{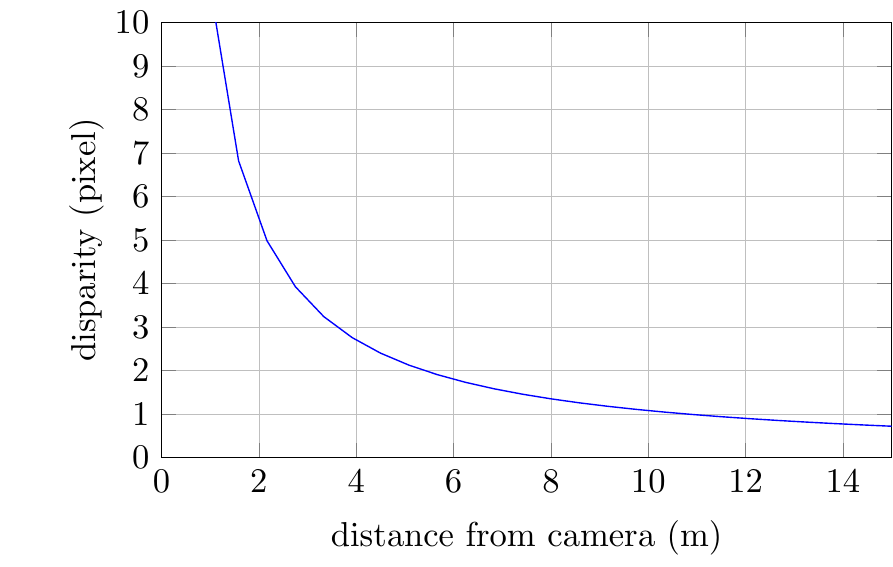}%
    \caption{The estimated theoretical disparity for the proposed setup.}%
    \label{fig:estimatedTheoreticalDisparity}%
\end{figure}

For the sake of safety, in the considered scenarios all the targets were observed from a minimum distance of \SI{2}{\meter} considering objects up to \SI{1}{\meter} wide. Starting from these assumptions, the maximum baseline length $b_\mathrm{max}$ can be retrieved by inverting eqs.~\eqref{eq:mindist} and~\eqref{eq:focalLength} and assuming that the flat $W =$ \SI{1}{\meter} wide, the target distance is $z_m =$ \SI{2}{\meter} along the optical lines, and the focal length is $f \approx 54~\text{pixels}$, as follows:
\begin{equation}
  b_\mathrm{max} = 2\cdot \tan(\xi/2)\cdot \left(z_m - \frac{w_o/2}{\tan(\xi/2)}\right) \approx 0.2~\text{m}.
\end{equation}
Figure~\ref{fig:estimatedTheoreticalDisparity} shows the estimated theoretical disparity (i.e., $d=x_r - x_l$) for the proposed hardware setup (see Sec.~\ref{sec:systemOverview}). As can be seen from the graph, the larger the distance from the camera is, the lower the disparity is. Hence, the greater the difference with the ground truth value.








\section{Experimental verification}
\label{sec:experimentalVerification}

To demonstrate the feasibility and the effectiveness of the proposed hardware and software setup, numerical simulations in Gazebo and real flight experiments were carried out. In particular, the Gazebo robotics simulator was used to simulate the system in a realistic scenario, exploiting the advantages of Software-in-the-loop simulations~\cite{SilanoSMC19}. In addition, Gazebo simulations were used to reduce the probability of failures and to obtain a qualitative analysis of the algorithms described in the previous sections. At this stage, the vehicle dynamics is not considered. A Gazebo plugin simulates the sensor model in such a way that the environment realistically reflects the actual world. Real flight tests verified not only the accomplishment of the assigned mission (i.e., the detection of overheated objects and the recovery of their relative position~\ac{wrt} the vehicle position), but also the reliability of the hardware setup. All the simulations were performed on a laptop with an i7-8565U processor (1.80GHz) and 32GB of RAM running on Ubuntu 18.04.





\subsection{Simulation results}
\label{sec:simulationExperiments}

\begin{figure}
  \centering
  \includegraphics[trim={0pt 0pt 0pt 80pt},clip,width=0.93\linewidth]{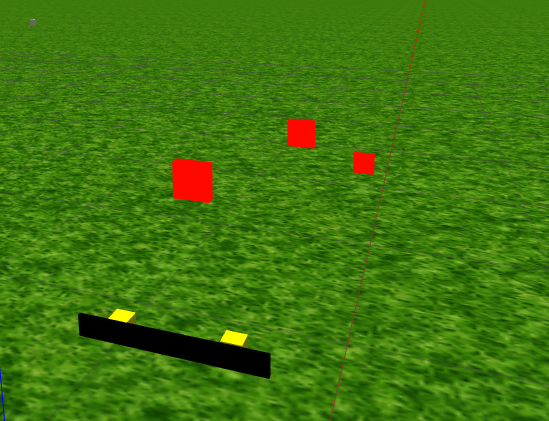}%
  \caption{Gazebo simulation setup in the case of three heat elements in front of the cameras.}
  \label{fig:simulation_setup}
\end{figure}

The Gazebo simulation setup consists of two cameras (yellow) on a holder (black) and several heat elements (red) located at various distances from the camera, as depicted in Fig.~\ref{fig:simulation_setup}. Both in simulation and real flight experiments the baseline was set to \SI{0.2}{\meter}, following the 3D localization considerations detailed in Sec.~\ref{sec:stereoGeometry}.


\begin{figure}
  \begin{center}
    \includegraphics[width=0.9\columnwidth, trim={0pt 0pt 0pt 0pt},clip]{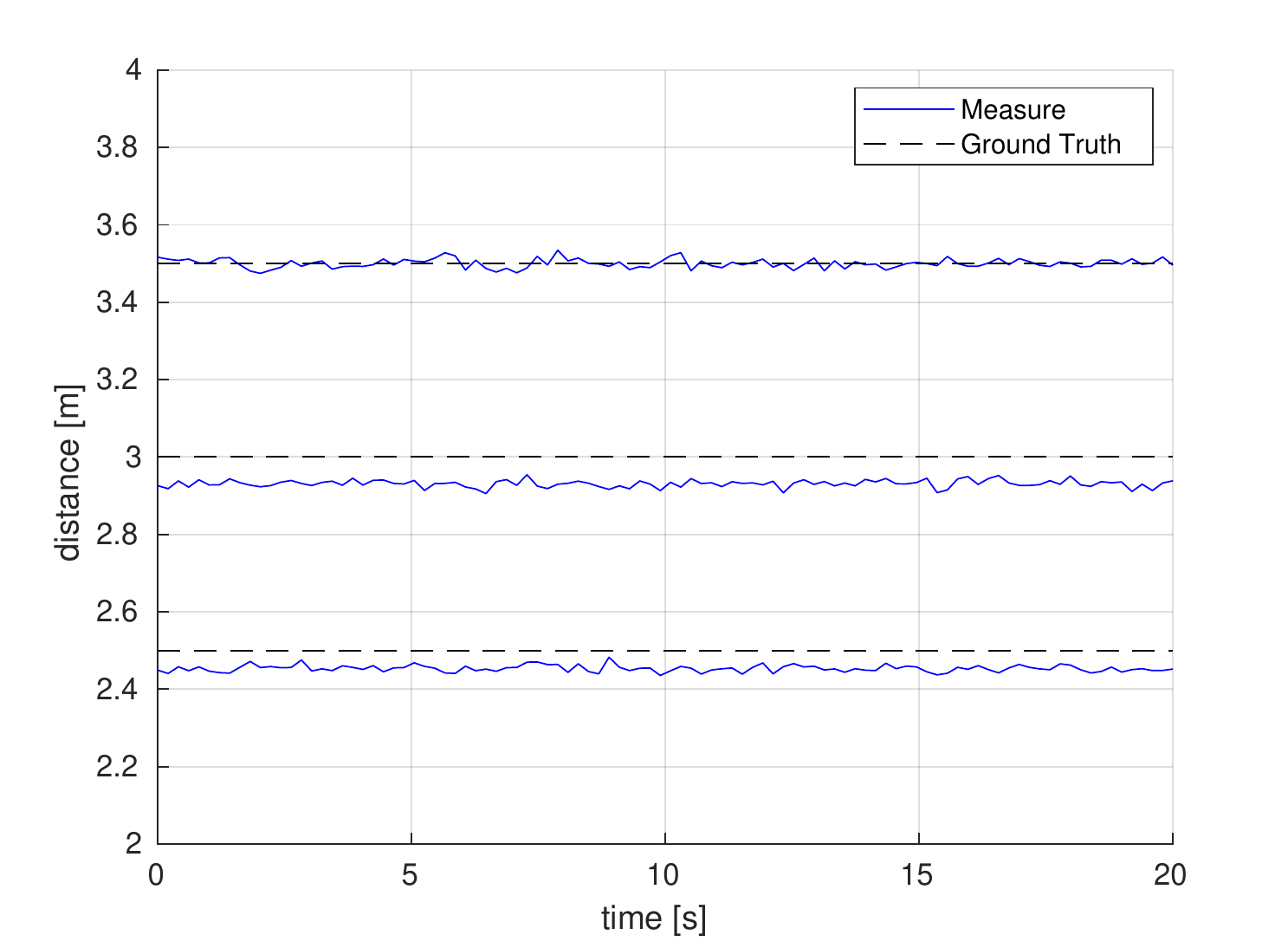}
   \end{center}
  \vspace{-0.75em}
  \caption{Simulation results in Gazebo in the case with two heat elements. Solid lines represent the sensor measurements, while dashed lines are the ground truth values.
  }
  \label{fig:sim_2_obj}
\end{figure}

\begin{figure}
  \begin{center}
    \includegraphics[width=0.9\columnwidth, trim={0pt 0pt 0pt 0pt},clip]{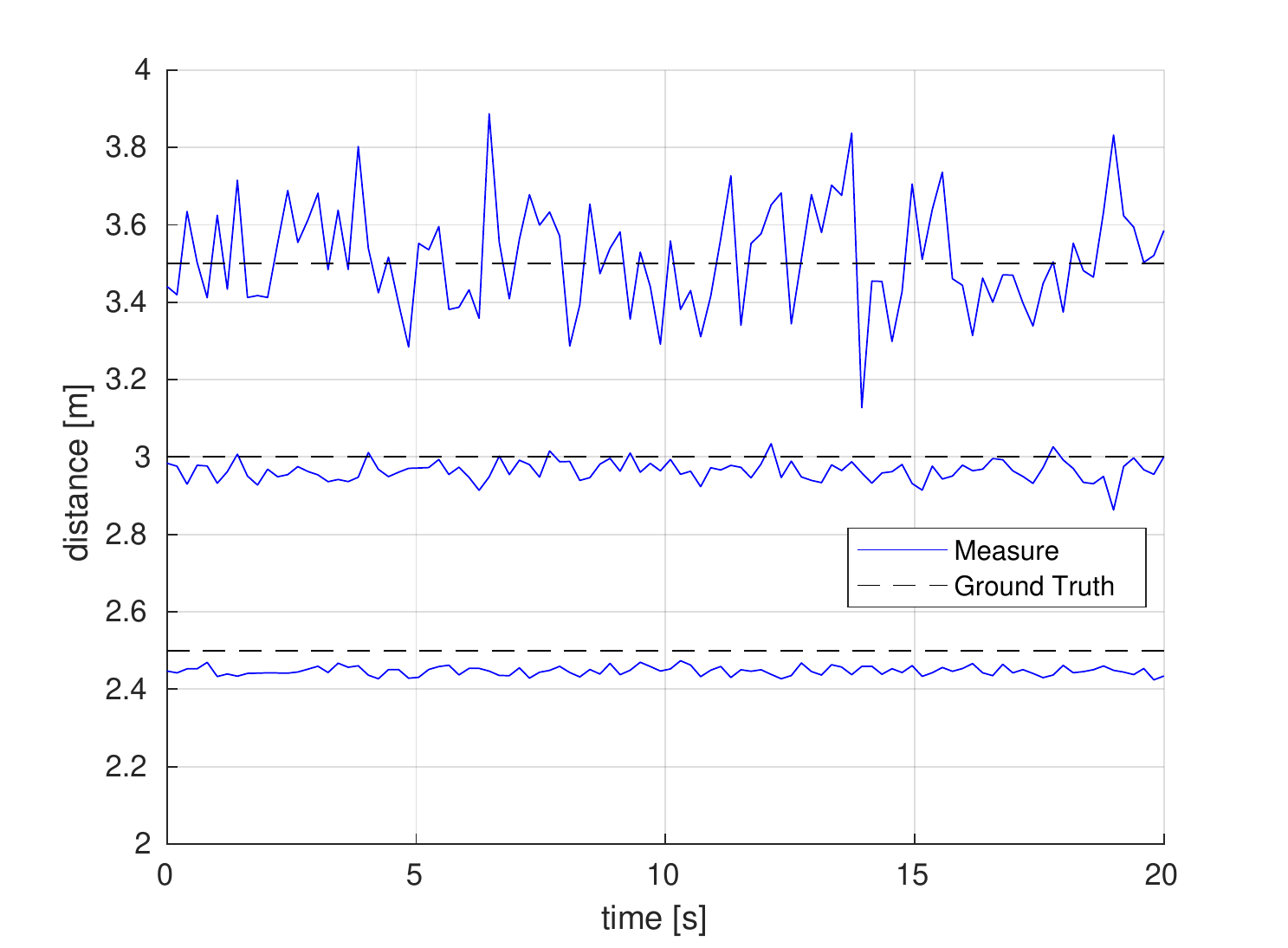}
  \end{center}
  \vspace{-0.75em}
  \caption{Simulation results in Gazebo in the case with three heat elements.
  }
  \label{fig:sim_3_obj}
\end{figure}

For ease of experimentation, only two testbed scenarios were considered, i.e., the case with two (see Fig.~\ref{fig:sim_2_obj}) and three heat objects (see Fig.~\ref{fig:sim_3_obj}) in the environment, but this does not imply a loss of generality of the approach. Real flight experiment were performed to demonstrate the validity of the proposed solution, as detailed in Sec.~\ref{sec:realWorldExperiments}. In both scenarios, the task objective was to measure the relative position of a target heat object~\ac{wrt} the vehicle position. Environmental perturbations were modeled by including several heat objects at various temperatures (above \SI{125}{\degree} to avoid incurring in outlier detection, as described in Sec.~\ref{sec:vision}) in the simulation which helped yield unbiased estimates of the relative position. A summary of the effects of measurement errors and environmental perturbations on the relative position estimates is shown in Figs.~\ref{fig:sim_2_obj} and~\ref{fig:sim_3_obj}.

It is worth noticing that when environmental perturbations are large, estimates of the position are likely to be reliable even when measurement errors are also large. By contrast where the environment is relatively constant, unbiased estimates of the position can only be obtained if populations are counted precisely. 

As can be seen from the graph, the sensing capabilities depend both on the distance and on the number of the heat element: whether it is close to position relative to the stereoscopic axis or distinct from it. 

A quantitative analysis of the depth estimation was carried out by varying the distance of the target in the range $2$ to \SI{3.5}{\meter}, with increments of \SI{0.5}{\meter} in position, and measuring the mean and the standard deviation of the collected measurements, as detailed in Table~\ref{tab:sim_2}. For all the experiments, the measuring time was maintained constant and equal to \SI{20}{\second}. A safety distance of \SI{2}{\meter} guarantees the compliance with safety requirements between drones and heat object while ensuring that the mission is carried out successfully.


\begin{table}[ht]
  \begin{center}
    \begin{tabular}{|l|c|c|c|}
    \hline
    & \textbf{Object Distance} & \textbf{Mean} & \textbf{Std Deviation} \\
    \hline \hline
    \multirow{3}{*}{Two heat objects} & \SI{2.5}{\meter} & \SI{2.454}{\meter} & \SI{0.009}{\meter} \\
    & \SI{3.0}{\meter} & \SI{2.931}{\meter} & \SI{0.009}{\meter} \\
    & \SI{3.5}{\meter} & \SI{3.502}{\meter} & \SI{0.012}{\meter} \\
    \hline \hline
    \multirow{3}{*}{Three heat objects} & \SI{2.5}{\meter} & \SI{2.447}{\meter} & \SI{0.011}{\meter} \\
    & \SI{3.0}{\meter} & \SI{2.964}{\meter} & \SI{0.027}{\meter} \\
    & \SI{3.5}{\meter} & \SI{3.466}{\meter} & \SI{0.143}{\meter} \\
 \hline
    \end{tabular}
  \end{center}
  \caption{Mean and standard deviation in the experiments with two and three heat elements in Gazebo.}
  \label{tab:sim_2}
\end{table}




\subsection{Real flight experiments}
\label{sec:realWorldExperiments}

\begin{figure}
  \begin{center}
    \includegraphics[width=0.95\columnwidth, trim={0pt 0pt 0pt 0pt},clip]{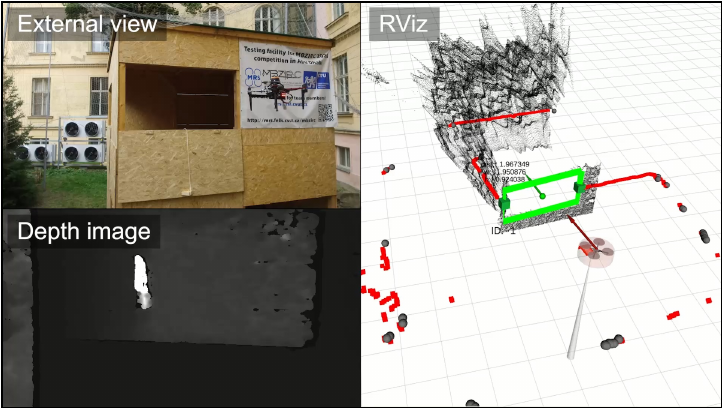}
   \end{center}
  \caption{View from an external camera, depth image, and RViz visualization of the real flight experiments. The visualization contains LiDAR data shown as red squares.
  }
  \label{fig:experimentsPhoto}
\end{figure}

\begin{figure}
  \begin{center}
    \includegraphics[width=0.9\columnwidth, trim={0pt 0pt 0pt 0pt},clip]{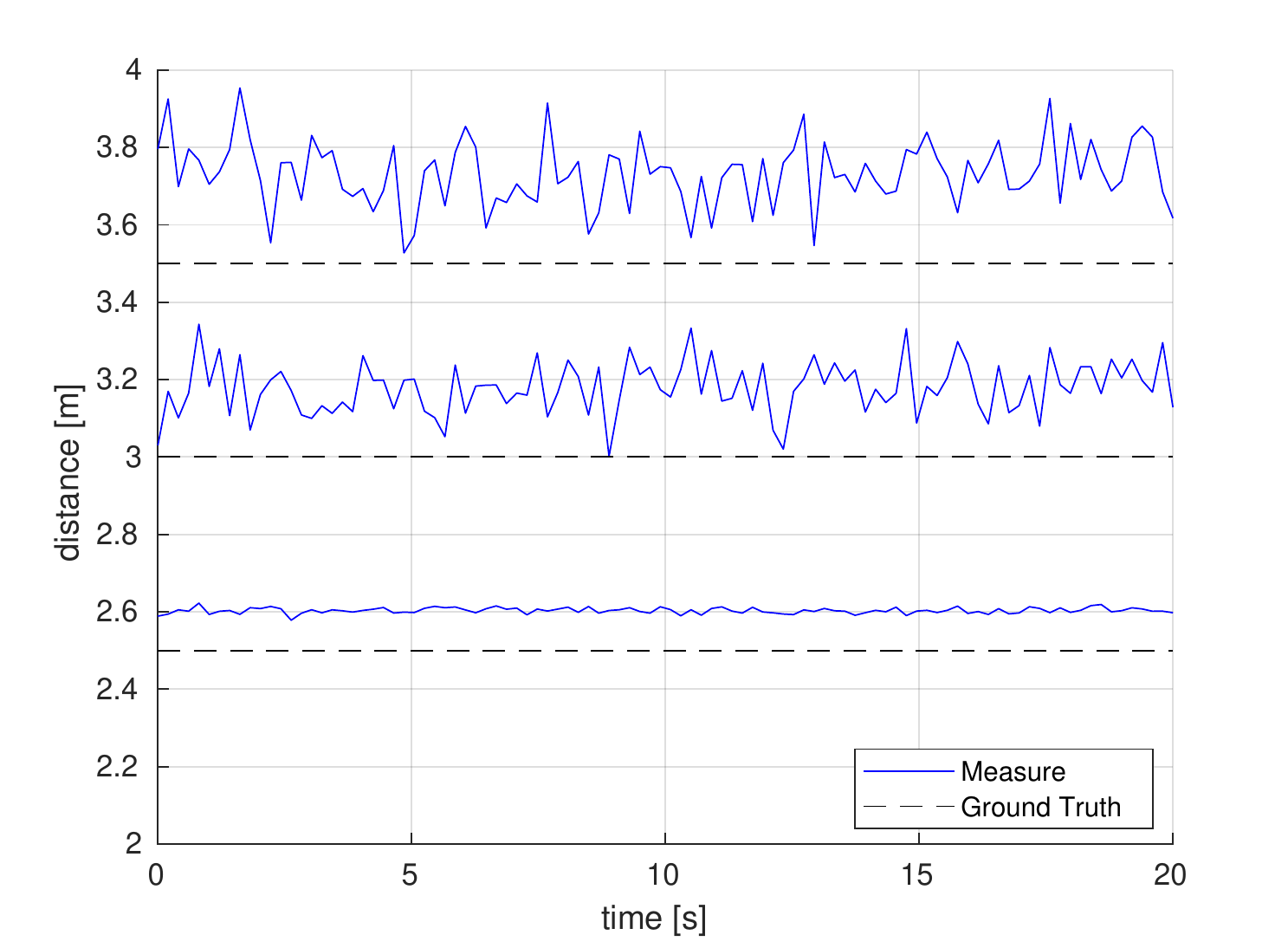}
  \end{center}
  \vspace{-0.75em}
  \caption{Simulation results in real flight experiments with three heat elements in front of the cameras.
  }
  \label{fig:real_3_obj}
\end{figure}

\begin{figure}
  \begin{center}
    \includegraphics[width=0.9\columnwidth, trim={0pt 0pt 0pt 0pt},clip]{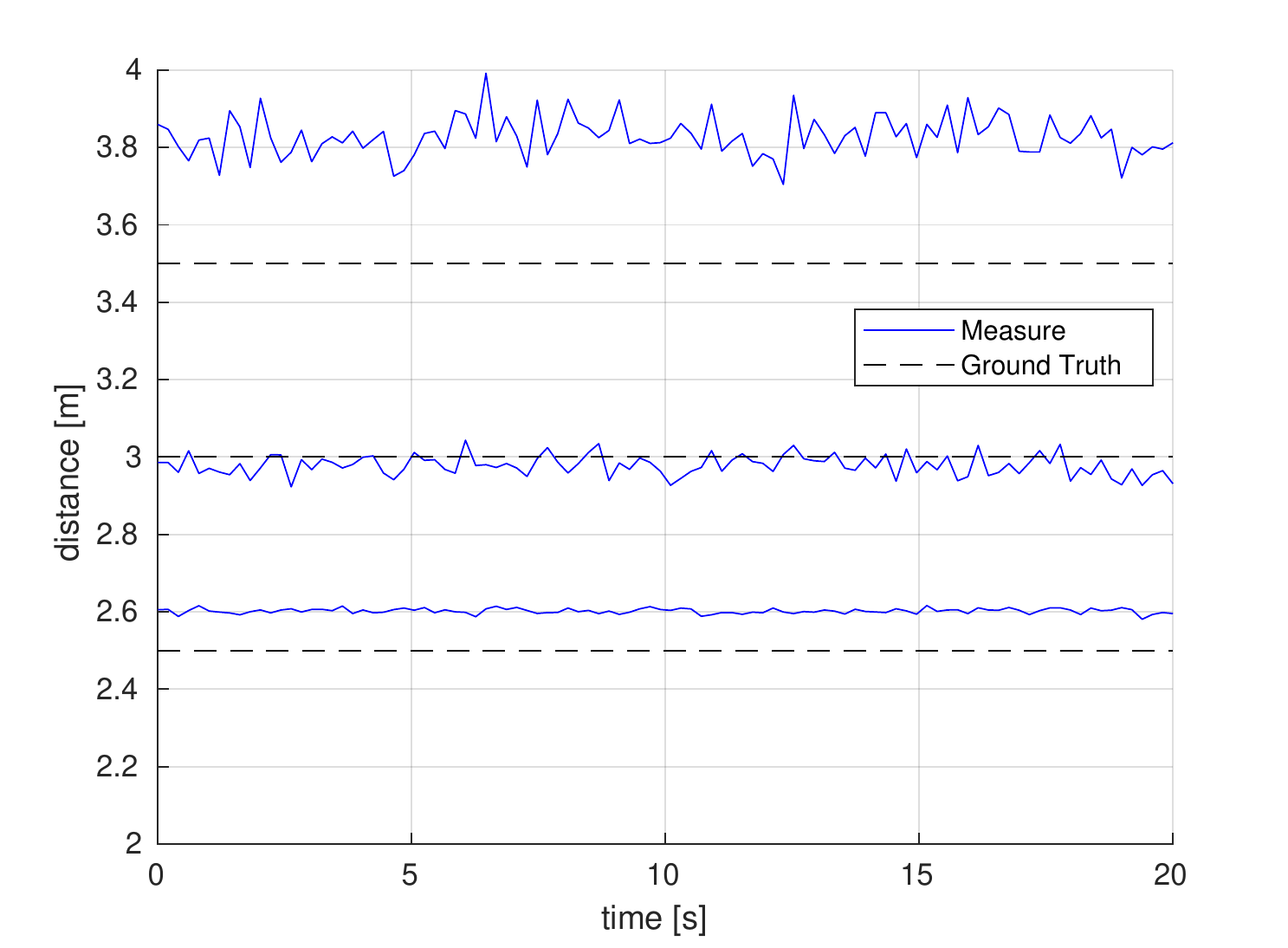}
   \end{center}
  \vspace{-0.75em}
  \caption{Simulation results in real flight experiments with two heat elements in front of the cameras.
  }
  \label{fig:real_2_obj}
\end{figure}

To evaluate and prove the applicability of the proposed hardware and software setup in real-world inspection tasks, real flight experiments were performed, as shown in Fig.~\ref{fig:experimentsPhoto}. As for the Gazebo simulations, two testbed scenarios were considered using two (see Fig.~\ref{fig:real_2_obj}) and three heat objects (see Fig.~\ref{fig:real_3_obj}). 

Compared to the Gazebo simulations, the inclusion of the drone dynamics in the loop enlarged the measurements errors, and thus the mean and the standard deviation values, as detailed in Table~\ref{tab:experimentData}. However, good results still confirmed the validity of the architecture while also highlighting the drawbacks and limitations of the proposed solution: the low-resolution of the cameras compared to their~\ac{FOV}, as well as the typical searching distance of approximately \SI{2}{\meter}~\ac{wrt} the vehicle, made it more complicated to detect heating elements. 

For instance, objects with a temperature of~\SI{150}{\degree} could be parsed by the cameras as having approximately only~\SI{108}{\degree}. This effect further decreased the temperature read out of the pixels containing the heating elements, as the camera effectively averages the temperatures of the entire surface captured within a single pixel. In relatively cold and shaded surroundings, these elements may be distinguishable as concentrated patches of elevated temperature. However, in the real-world conditions of the experiments these were barely noticeable, which would not be the case for real maintenance operations. 

As depicted in Figs.~\ref{fig:sim_3_obj} and~\ref{fig:real_3_obj}, the larger the distance the between the drone and heat object is, the lower the accuracy of the distance estimation is. The low-camera resolution has a significant impact on the distortion of measurement data. Thus, the lower the resolution, the greater the total projected area per pixel, and the less precise the reading will be.

For all such reasons, the platform should be meant as a tool to perform preliminary inspection operations of electrical power lines. 

\begin{table}[ht]
  \begin{center}
    \begin{tabular}{|l|c|c|c|}
    \hline
    & \textbf{Object Distance} & \textbf{Mean} & \textbf{Std Deviation} \\
    \hline \hline
    \multirow{3}{*}{Two heat objects} & \SI{2.5}{\meter} & \SI{2.603}{\meter} & \SI{0.006}{\meter} \\
    & \SI{3.0}{\meter} & \SI{3.981}{\meter} & \SI{0.030}{\meter} \\
    & \SI{3.5}{\meter} & \SI{3.832}{\meter} & \SI{0.048}{\meter} \\
    \hline \hline
    \multirow{3}{*}{Three heat objects} & \SI{2.5}{\meter} & \SI{2.603}{\meter} & \SI{0.007}{\meter} \\
    & \SI{3.0}{\meter} & \SI{3.180}{\meter} & \SI{0.066}{\meter} \\
    & \SI{3.5}{\meter} & \SI{3.737}{\meter} & \SI{0.093}{\meter} \\
 \hline
    \end{tabular}
  \end{center}
  \caption{Mean and standard deviation in the experiments with two and three heat elements in real flight experiments.}
  \label{tab:experimentData}
\end{table}

\section{Conclusions}
\label{sec:Conclusions}

This paper has presented an application of stereo thermal vision for preliminary inspection operations of electrical power lines. In particular, a hardware and software setup were described to allow the detection of overheated power equipment by finely detecting the potential source of damage while also providing a measure of the harm extension. The architecture is designed around a set of software routines that handle the current states of the system ensuring compliance with safety requirements. Numerical simulations in Gazebo and real flight experiments demonstrated the validity and the effectiveness of the proposed setup, showing its applicability in real world scenarios. Future work includes investigating better computer vision solution to deal with the correspondence matching problem and uncertainties and thermal noise in task execution.

\bibliographystyle{IEEEtran}
\bibliography{bib.bib}

\end{document}